\documentclass[a4paper,twocolumn,twoside,10pt]{ranlp}

\usepackage{epsfig}

\usepackage[usenames]{color}

\newcommand{\IMDB}{\textsc{imdb}}
\newcommand{\POS}{\textsc{pos}}
\newcommand{\LM}{\textsc{lm}}

\DeclareMathOperator*{\argmax}{arg\,max}

\begin{document}

\title{\textbf{Language-Independent Sentiment Analysis\\
Using Subjectivity and Positional Information}}

\author{
Veselin Raychev and Preslav Nakov\footnote{Also: Department of Computer Science,
National University of Singapore, 13 Computing Drive, Singapore 117417, nakov@comp.nus.edu.sg}\\
Department of Mathematics and Informatics\\
Sofia University ``St Kliment Ohridski''\\
5, James Bourchier Blvd.,1164 Sofia, Bulgaria\\
\email{\{veselin.raychev, preslav.nakov\}}{fmi.uni-sofia.bg}
}

\date{} 
\maketitle 

\thispagestyle{empty}	     \pagestyle{empty}

\begin{abstract}
We describe a novel language-independent approach to the task of
determining the polarity, positive or negative, of the author's opinion
on a specific topic in natural language text.
In particular, weights are assigned to attributes, individual words or word bi-grams,
based on their position and on their likelihood of being subjective.
The subjectivity of each attribute is estimated in a two-step process,
where first the probability of being subjective is calculated
for each sentence containing the attribute,
and then these probabilities are used to alter the attribute's weights
for polarity classification. The evaluation results on a standard dataset
of movie reviews shows 89.85\% classification accuracy,
which rivals the best previously published results for this dataset for systems
that use no additional linguistic information nor external resources.
\end{abstract}

\keywords{Sentiment analysis, subjectivity identification, polarity classification, text categorization.}


\section{Introduction}

Recently, there has been growing research interest
in determining the polarity, positive or negative,
of the author's opinion on a specific topic in natural language texts.
Such analysis has various potential applications
ranging from components for web sites
to business and government intelligence \cite{Pang:Lee:2008}.
Previous research on document sentiment classification
has shown that machine learning based classifiers
perform much better compared to rule-based systems \cite{pang:lee:vaithyanathan:2002}.
However, the task remains challenging
since opinions are typically expressed in a specific manner,
using many rare words and language expressions.
As previous research has shown \cite{Wiebe:al:2004},
even words with a single occurrence on training
can turn out to be good predictors on testing.
As a result, the classification accuracy for sentiment analysis
using machine learning approaches tends to be much lower
compared to that for other text classification tasks
like topic identification.


\section{Related work}

Pang \& al. \cite{pang:lee:vaithyanathan:2002} pioneered the field of sentiment analysis.
They worked on a \emph{sentiment polarity classification} task,
choosing between a positive and negative label
using Na\"{i}ve Bayes and support vector machines (\SVM),
where each text document was represented as a bag-of-words with weights for word presence.
They further tried to use negation, word positions and part-of-speech (\POS) information
without much success, and found that many techniques that typically
help for topic classification negatively affected the accuracy for sentiment polarity.
The experiments were carried out on a set of 2,000 movie reviews mined from the web,
1,000 positive and 1,000 negative,
without explicit information about polarity,
i.e., without ranks, scores, or number of stars.
The dataset was made publicly available\footnote{http://www.cs.cornell.edu/people/pabo/movie-review-data}
and has since become the de-facto standard for training and evaluation
in most of the subsequent research.

In the case of movie reviews,
sentiment polarity classification has been found to be hard
not only because of many informative words being rare,
but also due to large portions of the movie reviews consisting
of non-subjective sentences that just narrate the movie plot
without actually contributing much sentiment information.
In an attempt to get rid of such sentences,
Pang and Lee \cite{pang-lee:2004:ACL}
proposed a pre-processing filter that removes all non-subjective sentences
while retaining the subjective ones to be used for sentiment polarity classification.
In order to train that filter, they created a special dataset consisting
of 5,000 subjective and 5,000 non-subjective sentences mined from the
\emph{Internet Movie DataBase}\footnote{http://www.imdb.com} (\IMDB).
This gave rise to a new task, \emph{subjectivity classification},
as an intermediate step for polarity classification.
In their experiments, Pang and Lee used a Na\"{i}ve Bayes classifier,
which yielded 92\% accuracy for the subjectivity filter.
Using the filter to help choose subjective sentences for polarity classification
yielded 86.4\% accuracy, which represents about 3\% absolute improvement
for the sentiment polarity classification with a Na\"{i}ve Bayes classifier;
there were no improvements when using an \SVM\ classifier.

Matsumoto \& al. \cite{Matsumoto:Takamura:Okumura:05a}
experimented with an \SVM\ classifier and a more recent version
of the polarity dataset.
Using several innovative features based on linguistic analysis,
including unigrams, bigrams and all pairs of words within the same sentence,
they achieved over 88.1\% accuracy when only language-independent features were used,
and 92\% when additional English-specific linguistic information was introduced.

There have been some attempts to use language models (\LM)
for polarity classification, but the resulting accuracy was low.
Hu \& al. \cite{Hu:al:2007} tried using language models (\LM) for polarity classification
with several different kinds of smoothing,
but found that a model based on unigrams, i.e., without sequence information,
performed better.
One possible explanation could be found in the observation that,
for the task of sentiment polarity classification,
the Na\"{i}ve Bayes classifier works better when the feature weights
are binary (i.e., when only term presence/absence is taken into account,
but repetitions are ignored) rather than frequency-based,
while language models calculate the probability to generate a document
taking term repetitions into account.

Below we propose a novel approach that assigns weights to individual attributes,
words or word bi-grams, based on their position in the text
and on their likelihood of being subjective.
Using the Na\"{i}ve Bayes classifier, we achieve 89.85\% accuracy,
which is an improvement over the best previously published
language-independent results
that use no additional linguistic information sources
such as parsers, POS taggers, stemmers, etc.


\section{Method}

In this section,
we first describe the multinomial Na\"{i}ve Bayes classifier
and the way we are changing it.
We then explain how we use the subjectivity dataset
to improve the results further.

\subsection{Na\"{i}ve Bayes}

We use the Na\"{i}ve Bayes multinomial classifier,
which makes the na\"{i}ve assumption
that the occurrences of the attributes
(in our case: words and word bigrams) in a document
are conditionally independent given the document class
(in our case: `positive' or `negative').
It further assumes that the occurrences of the attributes
are position- and context-independent,
and that the document length is class-independent.
Each document is represented as a vector of attribute counts $x$
and its class-conditional probability is given by a multinomial distribution
over the set of attributes:

\begin{equation}
\mathrm{Pr}(x|c) = \mathrm{Pr}(l_x) \frac{l_x!}{\prod_{d}{x_d!}} \prod_{d}{\mathrm{Pr}(d|c)^{x_d}}
\end{equation}

\noindent where $l_x$ denotes the length of document $x$,
$c$ is a candidate class,
$d$ ranges over the set of all attributes occurring in document $x$,
and $x_d$ is the occurrence frequency of attribute $d$ in document $x$.

Using the Bayes rule, we can express the posterior probability
for class $c$ given document $x$ as follows:

\begin{equation}
\mathrm{Pr}(c|x) = \frac{\mathrm{Pr}(c) \prod_{d}{\mathrm{Pr}(d|c)^{x_d}}} {\sum_{c^{\prime}}{\mathrm{Pr}(c^{\prime}) \prod_{d}{\mathrm{Pr}(d|c^{\prime})^{x_d}}}}
\end{equation}

Then, the most likely class $\hat{c}$ for a document $x$
is selected as follows:

\begin{equation}\label{eq:mle}
\hat{c} = \argmax_{c}{\mathrm{Pr}(c|x)}
\end{equation}

After removing the denominator, which is independent of $c$,
and after taking a logarithm,
we obtain the following formula for the classification decision:

\begin{equation}\label{eq:hat}
\hat{c} = \argmax_{c}{[\log \mathrm{Pr}(c) + \sum_{d}{x_d \log \mathrm{Pr}(d|c)}]}
\end{equation}

Let $N_{cd}$ be the sum of the values $x_d$ of all attributes $d$
that occur in training documents $x$ that belong to class $c$:

\begin{equation}
N_{cd}= \sum_{x:class(x)=c}{x_d}
\end{equation}

Then the conditional probabilities $\mathrm{Pr}(d|c)$ can be estimated as follows:

\begin{equation}\label{eq:nosmoothing}
\mathrm{Pr}(d|c) = \frac{N_{cd}}{\sum_{d^{\prime}}{N_{cd^{\prime}}}}
\end{equation}

In order to avoid zero-valued estimates of attribute values,
the above probability should be smoothed \cite{Juan:Ney:2002}.
In our experiments, we use Laplace smoothing,
which estimates $\mathrm{Pr}(d|c)$ as follows:

\begin{equation}\label{eq:smoothing}
\mathrm{Pr}(d|c) = \frac{N_{cd} + s}{\sum_{d^{\prime}}{(N_{cd^{\prime}} + s)}}
\end{equation}

We set the smoothing parameter $s$ to 1, which is a commonly used default value.

\subsection{Positional Information}
\label{sec:pos}

The above-described multinomial Na\"{i}ve Bayes model
does not take into account the position of occurrence
of the attributes: for topic categorization tasks,
the occurrence frequency $x_d$ of attribute $d$ in document $x$
is typically used as a feature weight,
in the multinomial Na\"{i}ve Bayes model
and for sentiment polarity classification,
binary attributes for word presence
have been reported to yield
better classification accuracy \cite{pang:lee:vaithyanathan:2002}.
Still, in both cases, no positional information is being used.

In the above description,
each occurrence of attribute $d$ in document $x$
would contribute a count of 1 to the frequency $x_d$
regardless of the position it occurs at.
However, position seems to be playing an important role
since opinions in movie reviews tend to be expressed around the end of the document.
In order to account for this observation,
we introduce a new schema, where instead of 1,
an occurrence of attribute $d$ in document $x$
contributes a different value to the frequency $x_d$
depending on its position in $x$:
an attribute starting at position 0 counts as some constant $a$, $a \geq 0$,
and one starting at the last word in the document counts as $b=a+q$, $q > 0$.
Attributes occurring in between get position-dependent fractional counts
that are obtained using a simple linear interpolation,
namely $a + q \times \frac{p}{|x| - 1}$,
where $p$ is the position of occurrence of the attribute
and $|x|$ is the length of document $x$ in words.

Consider, for example, the following sample document (tokenized and lowercased):\\

\noindent \texttt{i have to admit that i was a little skeptical
as to how much I could really get out of another " anti-slavery " movie .
fortunately , i turned out to be wrong .}\\

The attribute for the word \emph{have}
occurs at position 1 and thus will get a fractional count of $a + q \times \frac{1}{34}$;
this will be also the value of its $x_d$.
Similarly, the attribute for the bigram \emph{be wrong} occurs at position 32,
and thus its $x_d$ will be $a + q \times \frac{32}{34}$.
Finally, the attribute for the word \emph{to} occurs three times,
at positions 2, 11, and 31, which count as $a + q \times \frac{2}{34}$,
$a + q \times \frac{11}{34}$, and $a + q \times \frac{31}{34}$, respectively;
the corresponding weight $x_d$ should be the sum of the three fractional counts.
However, since we are interested in sentiment polarity classification,
where binary attributes for word presence work better,
we will only take into account the last occurrence of the attribute
and thus we will set the value of $x_d$ to be the fractional count for that last occurrence.

Let us now see how using such fractional counts impacts the classifier.
First, let $a$ be 0.
According to eq.~(\ref{eq:nosmoothing}),
the conditional probability $\mathrm{Pr}(d|c)$
will be independent of the value of the parameter $q$;
however, as eq.~(\ref{eq:smoothing}) shows,
this will not be the case if smoothing is being used.
Let $a \neq 0$: then the fractional counts
are in the interval [$a$;$a+q$],
which can be seen as a scaled version
of the interval [$1$;$1+q^{\prime}$],
where $q^{\prime} = q/a$.
Now, let us further take into account the fact that in the movie reviews dataset
there is an equal number of positive and the negative reviews.
Then, we can rewrite eq.~(\ref{eq:hat}) as follows:

\begin{equation}\label{eq:hatnew}
\hat{c} = \argmax_{c}{\sum_{d}{x_d \log \mathrm{Pr}(d|c)}}
\end{equation}

From the last equation, we can see that,
if we multiply the values of all attributes by the constant $a$, $a \neq 0$,
the classification decision will remain the same (provided that we use no smoothing).

Thus, it is enough to consider two groups of classifiers,
\texttt{0+q} and \texttt{1+q}.
For \texttt{0+q}, the classifiers are equivalent for all values of $q$
(except for smoothing), which means that it is enough to test with $q = 1$.
Note that changing $q$ would be equivalent to updating the smoothing parameter $s$
for Laplace smoothing.

For comparison purposes,
we also apply a simpler scheme where we remove all attributes
that appear at the first $k$ positions in the document,
assuming they contribute no sentiment information.
This is similar to the approach adopted by Pang and Lee \cite{pang-lee:2004:ACL},
where some of the objective sentences were filtered out.

\subsection{Subjectivity}
\label{sec:subj}

Pang and Lee \cite{pang-lee:2004:ACL} used a subjectivity filter
to eliminate the non-subjective sentences in a target movie review,
so that they could apply their polarity classifier
on a smaller set of higher-quality sentences.
Although 92\% accurate, their filter is not perfect,
which could result in some useful features being lost.
In contrast, our weighting scheme can benefit
from the potential subjectivity of the last sentences
while still giving some smaller weight to the words in the earlier sentences.

In order to further benefit from the position-dependent weights,
we propose to move the subjective sentences to the end of the document.
We thus train a Na\"{i}ve Bayes classifier on the subjectivity dataset,
and we use its posteriors to estimate the likelihood
of each sentence being subjective;
we then use this likelihood to sort the sentences in decreasing order.

A potential drawback of this approach is that,
if all sentences turned out to be subjective,
it would be unable to take this into account.
This could be addressed by combining the approach
with non-subjective sentence filtering:
if we only sort sentences according to subjectivity,
\texttt{0+q} methods should perform well,
while when we also use filtering, \texttt{1+q} methods should be better
since the first subjective sentences
would get a high positive weight rather than one close to 0.


\section{Experiments and evaluation}

In our experiments, we used the above-described sentiment polarity dataset.
Unfortunately, it is not divided into proper training and testing subsets,
and thus we were forced to use a 10-fold cross-validation
in order to be directly comparable to previous publications.

However, there are some complications
since we further want to be able to optimize some parameters
such as the value of $q$.
Normally, this requires having three separate datasets:
(1) training,
(2) development,
and (3) testing.
In order to obtain a development dataset,
for each iteration of the 10-fold cross-validation,
we further perform an internal 5-fold cross-validation
which divides the training dataset into a train-train and a train-dev datasets:
the former is used for training the classifier,
while the latter is used for tuning the additional parameters.
After having chosen the values for the parameters,
we can train on the full training dataset.

\begin{figure}[tbhp]
\epsfig{file=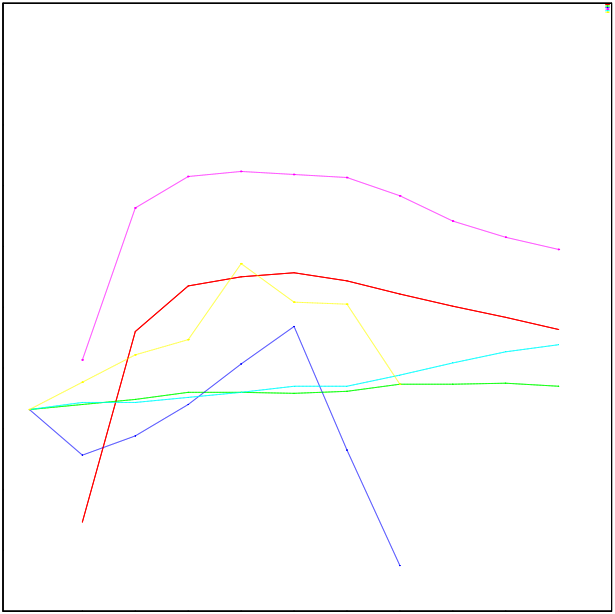, width=3.2in}
\caption{Accuracy with unigrams only:
         using filtering and sorting sentences by subjectivity.
         The horizontal axis shows the value of the parameter $q$.}
\label{fig:fig1}
\end{figure}

\begin{figure}[tbhp]
\epsfig{file=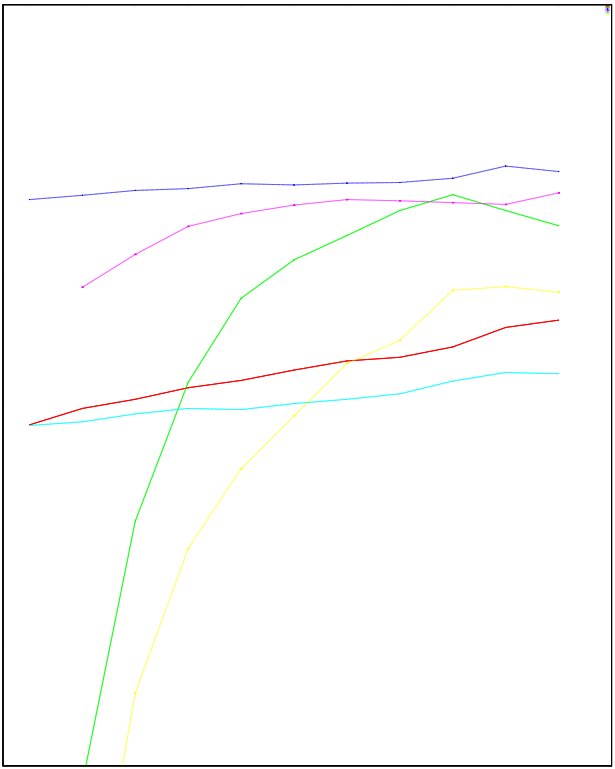, width=3.2in}
\caption{Accuracy with unigrams and bigrams:
         using filtering and sorting sentences by subjectivity.
         The horizontal axis shows the value of the parameter $q$.}
\label{fig:fig2}
\end{figure}

\subsection{Using unigrams only}

Unigrams, or just words, are the most widely used attributes in sentiment analysis,
and we show that the approaches proposed in sections \ref{sec:pos} and \ref{sec:subj}
do yield improvements in accuracy when used with unigrams.

Our baseline accuracy on the sentiment polarity dataset was 83.33\%
for the multinomial Na\"{i}ve Bayes classifier with Laplace smoothing.

Figure \ref{fig:fig1} shows that removing attributes
performs worse than updating their weights.
Fully removing the first few words in the documents yields a decrease
when done for the first 10\% of the words.
However, there are some benefits of having less noise,
e.g., when removing all words after the first 10\% up until 50\%.

The figure further shows that using \texttt{1+q} has little effect,
i.e., useless attributes are not penalized enough.
For \texttt{0+q}, the best result is achieved for $q=0.5$,
which yields 85.55\% accuracy with a corresponding
95\% Wilson confidence interval \cite{Agresti:Coull:1998:conf:interval}
of [83.94\%, 87.02\%];
it is a statistically significant improvement over the baseline.
However, the value of $q=0.5$ was chosen aposteriori,
and we need further verification to choose a value for $q$
based on the training dataset only.
Using the above-described nested cross-validation,
we achieved 86.42\% accuracy, which shows that the reported accuracy
was not due to the aposteriori selection.

We also experimented using the subjectivity dataset
to improve the accuracy of the classifier even further.
When we used filtering of the objective sentences as a baseline,
we achieved 86.31\% accuracy,
which is very close to what previous publications have reported
\cite{pang-lee:2004:ACL}.

All methods proposed in this paper for weighting attributes
yield improvements in accuracy when the sentences are sorted
according to subjectivity compared to when no sorting is used.
The \texttt{0+q} method with subjectivity sorting
achieves 87.23\% accuracy for $q = 0.4$.
Again, we need to prove that this value of $q$
does not yield a randomly good score
just because of the choice being made aposteriori.
Choosing a value based on the training dataset only,
using the nested 5-fold cross-validation,
yielded 87.62\% accuracy,
which means that it is very likely that the method
performs at least as good as the reported accuracy.

\begin{table*}[tbh]
\centering
{\small \begin{tabular}{lccc}
\textbf{Method} & \textbf{Accuracy} & \textbf{Reference} & \textbf{Subj.?}\\
\hline
Na\"{i}ve Bayes, unigrams & 83.33 & \cite{pang-lee:2004:ACL} & $-$\\
Unigrams, \texttt{0+q}, $q=0.5$ & 85.55 & this work & $-$\\
Na\"{i}ve Bayes, unigrams and bigrams & 85.59 & this work & $-$\\
Na\"{i}ve Bayes, unigrams, subjectivity filter & 86.40 & \cite{pang-lee:2004:ACL} & $+$ \\
Unigrams, \texttt{0+q}, $q=0.4$, subjectivity sort & 87.25 & this work & $+$ \\
Unigrams and bigrams, \texttt{0+q}, $q=1$ & 87.81 & this work & $-$ \\
\SVM, unigrams and bigrams & 88.10 & \cite{Matsumoto:Takamura:Okumura:05a} & $-$ \\
Na\"{i}ve Bayes, unigrams and bigrams, subjectivity filter & 89.30 & this work & $+$\\
Unigrams and bigrams, \texttt{0+q}, $q=1.5$, subjectivity sort and subjectivity filter & \textbf{89.85} & this work & $+$\\
\end{tabular}}
\caption{Comparing our results to those in previous publications
         using the sentiment polarity dataset: accuracy is shown in \%.
         The last column indicates whether the subjectivity dataset was used.}
\label{tab:comparison}
\end{table*}

\subsection{Using unigrams and bigrams}

A natural extension of the above methods is to add more features.
Previous research has shown that using different sets and methods
to add bigrams may improve or damage the accuracy of the classifier
\cite{Matsumoto:Takamura:Okumura:05a,pang:lee:vaithyanathan:2002}.
We show that adding bigrams improves the accuracy
when using the movie reviews dataset v2.0
with the full set of 1,000 positive and 1,000 negative documents.
Using unigrams and bigrams with the Na\"{i}ve Bayes multinomial classifier
yields 85.59\% accuracy, which is significantly better
than the accuracy of 83.33\% for unigrams only.
Similarly, when using the subjectivity dataset to filter objective sentences first,
unigram features yield 86.31\% accuracy
while using unigrams and bigrams together yields 89.30\% accuracy.

With position-dependent attribute weights,
we have three experimental conditions with respect to the subjectivity dataset:
(1) not using it,
(2) using it to sort sentences by subjectivity,
and (3) using it to filter the objective sentences
and then sort the remaining sentences by subjectivity.

The results are presented on Figure \ref{fig:fig2}.
Not using the dataset yields the maximum accuracy of 87.81\%
for the \texttt{0+q} method for $q=1$.
The corresponding 95\% Wilson confidence interval is [86.30\%, 89.17\%].
This is a statistically significant improvement
compared to the baseline, which does not use the subjectivity dataset: 85.59\% accuracy.

Using the subjectivity dataset allows for higher accuracy
to be achieved by the \texttt{0+q} method.
Sorting the sentences by subjectivity yields 89.38\% accuracy for $q = 1$.
However, this is not a statistically significant improvement
compared to the baseline that filters objective sentences.
Thus, the method does not perform that well with unigrams and bigrams
in combination with the subjectivity dataset.
The highest accuracy achieved by our methods is 89.85\%;
it is not statistically better than our baseline,
but still shows the potential of the method.


\section{Discussion}

The polarity classification task of movie reviews
has attracted a lot of research interest and many classifiers
have been applied to it so far.
As a result, support vector machines have been found
to be among the most accurate;
however, as Pang and Lee \cite{pang-lee:2004:ACL} have shown,
although the \SVM\ classifier perform very well on the polarity classification task,
removing subjective sentences fails to improve their accuracy.
Matsomoto \& al. \cite{Matsumoto:Takamura:Okumura:05a}
experimented with several methods to add different features
and reported that an \SVM\ classifier
with unigrams and bigrams yields 88.1\% accuracy.
Our best approach achieves 89.85\% accuracy
using multinomial Na\"{i}ve Bayes;
the corresponding 95\% Wilson confidence interval is [88.45\%, 91.10\%],
which makes it significantly better than the result for \SVM.
Note, however, that we are using the subjectivity dataset
in addition to the sentiment polarity one.

Language modeling represents another common approach to document classification.
Its popularity could be explained by its simplicity
and by the existence of several easy-to-use state-of-the-art implementations.
However, for polarity classification,
language modeling approaches generally perform poorly:
the best accuracy we could find is that of 
Hu \& al. \cite{Hu:al:2007}, who achieved a maximum accuracy of only 84.13\%.

Table \ref{tab:comparison} shows a summarized comparison
of the results from our experiments
with those reported in previous publications
using the sentiment polarity dataset.
The table also indicates which results have been obtained
using the subjectivity dataset as an additional dataset.


\section*{Acknowledgments}

The presented research is supported by the project FP7-REGPOT-2007-1 SISTER.


\section{Conclusion and future work}

We have described a novel approach to the task of
determining the polarity, positive or negative, of the author's opinion
on a specific topic in natural language text.
The approach uses language-independent features only
and makes no use of linguistic analysis.
The evaluation results on a standard dataset
of movie reviews have shown classification accuracy
that rivals the best previously published results
for this dataset for systems
that use no additional linguistic information nor external resources.

There are many ways in which the presented approach could be extended.
First, we would like to try combining our attribute weighting scheme
with more complex features such as subtrees of dependency trees,
as proposed by Matsumoto \& al. \cite{Matsumoto:Takamura:Okumura:05a};
note that this would make the resulting approach
dependent on a particular dependency parser,
thus yielding its language-independence questionable.
Another possible research direction would be
using an additional classifier such that,
given a list of the document sentences
sorted by the likelihood of being subjective in increasing order,
it can find the position after which all sentences are actually subjective;
they will be then given higher weights.
We would also like to experiment with other position-dependent
weighting functions, e.g., non-linear.
Using other classifiers is another interesting direction;
in particular, we are interested in finding a way to improve \SVM\
using the subjectivity dataset.
Finally, we plan to apply our approach to other domains and languages,
thus assessing the extent of validity of its underlying assumption.


\bibliographystyle{abbrv}  
\begin{scriptsize}
\bibliography{ranlp-biblio}  
\end{scriptsize}

\end{document}